# Systematic Exploration of Dialogue Summarization Approaches for Reproducibility, Comparative Assessment, and Methodological Innovations for Advancing Natural Language Processing in Abstractive Summarization

**Yugandhar Reddy Gogireddy[1]**

*dept.Computational Intelligence*

*SRM Institute of Science and Technology, Chennai*

ys0378@srmist.edu.in

**Jithendra Reddy Gogireddy[1]**

*dept.Computer Science and Engineering*

*SRM University, AP*

jithendra_gogireddy@srmap.edu.in

**ABSTRACT**

Reproducibility in scientific research, particularly within the realm of natural language processing (NLP), is essential for validating and verifying the robustness of experimental findings. This paper delves into the reproduction and evaluation of dialogue summarization models, focusing specifically on the discrepancies observed between original studies and our reproduction efforts. Dialogue summarization is a critical aspect of NLP, aiming to condense conversational content into concise and informative summaries, thus aiding in efficient information retrieval and decision-making processes. Our research involved a thorough examination of several dialogue summarization models using the AMI (Augmented Multi-party Interaction) dataset. The models assessed include Hierarchical Memory Networks (HMNet) and various versions of Pointer-Generator Networks (PGN), namely PGN(DKE), PGN(DRD), PGN(DTS), and PGN(DALL). The primary objective was to evaluate the informativeness and quality of the summaries generated by these models through human assessment, a method that introduces subjectivity and variability in the evaluation process. The analysis began with Dataset 1, where the sample standard deviation of 0.656 indicated a moderate dispersion of data points around the mean.

In our reproduction experiment, the scores were overall inferior to those in the original study, mostly all below 3.0 versus the original scores, which were all above 3.0. Specifically, the gold standard scores in our analysis were significantly lower than those reported in the original study. In our experiment, we noticed a distinct trend that contrasts with the findings of Feng et al. (2021), in which the performance of HMNet did not exhibit substantial gains over the PGN models. he original study regarding AMI/Informativeness did not demonstrate effectiveness: the baseline HMNet performed the best. This raises the question of whether the selection of the AMI dataset is appropriate for the human evaluation reproduction and the verification of the performance of the models using DialoGPT to achieve better performance in dialogue summarization. Furthermore, conducting a comprehensive analysis of dataset characteristics and evaluation metrics could offer valuable insights into enhancing the appropriateness of the dataset selection for evaluating summarization models.

## INTRODUCTION

The field of Natural Language Processing (NLP) has seen remarkable advancements in recent years, particularly in the area of dialogue summarization. Dialogue summarization is the process of condensing conversational content into concise, informative summaries. This capability is crucial for a wide range of applications, including customer service, meeting minutes, and information retrieval, where users need to quickly grasp the essential points from lengthy discussions. Despite the progress, the challenge of accurately summarizing dialogues remains significant due to the inherent complexity and dynamic nature of human conversations. Human conversations are intricate, involving multiple participants, varying topics, and numerous interruptions and overlaps. Unlike monologues or structured texts, dialogues are often informal and unstructured, making it challenging for summarization models to capture the essence effectively. The need for effective dialogue summarization tools is underscored by the increasing volume of conversational data generated across various domains, from business meetings to social media interactions. Automating the summarization process can save time and effort, improve information dissemination, and enhance decision-making processes. Several challenges characterize the task of dialogue summarization:

- **Complexity and Dynamics**: Dialogues are inherently more complex than monologues. They involve interactions among multiple speakers, leading to frequent topic shifts, interruptions, and overlapping speech, which complicates the summarization process.
- **Informality and Ambiguity**: Human conversations often include informal language, slang, and colloquialisms. Additionally, participants may leave sentences unfinished, use ambiguous references, or assume shared knowledge, making it difficult for models to generate accurate summaries.
- **Relevance and Redundancy**: Identifying the most relevant information and avoiding redundancy are critical in summarization. In dialogues, important points can be dispersed throughout the conversation, requiring the model to effectively filter and condense the content without losing essential information.

- **Evaluation Metrics**: Evaluating the quality of generated summaries is another challenge. Traditional metrics used for text summarization may not fully capture the nuances of dialogue summarization. Human evaluation, while insightful, introduces subjectivity and variability, necessitating more robust and standardized evaluation methods.

This study aims to address these challenges by reproducing and evaluating dialogue summarization models through human assessment. The focus is on assessing the informativeness and quality of summaries generated by different models using the AMI (Augmented Multi-party Interaction) dataset.

The specific objectives include:

- **Reproduction of Existing Studies**: To reproduce the results of existing studies on dialogue summarization models, particularly those involving Hierarchical Memory Networks (HMNet) and Pointer-Generator Networks (PGN), to validate their findings.
- **Evaluation of Model Performance**: To evaluate the performance of these models in generating informative and high-quality summaries through human assessment, identifying the strengths and weaknesses of each model.
- **Analysis of Variability and Consistency**: To analyze the variability and consistency in human evaluations, exploring the factors that contribute to discrepancies in the assessment of dialogue summaries.

Our approach involves a systematic reproduction of existing studies on dialogue summarization models, followed by a thorough evaluation of the generated summaries. The key steps in our methodology include:

- **Dataset Selection**: The AMI dataset, which consists of multi-party meeting recordings, was chosen for this study. This dataset provides a rich source of conversational data with varying levels of complexity and formality, making it ideal for evaluating dialogue summarization models.
- **Model Implementation**: Several dialogue summarization models were implemented, including HMNet and various versions of PGN (PGN(DKE), PGN(DRD), PGN(DTS), and PGN(DALL)). These models represent different approaches to capturing and condensing dialogue content.
- **Human Evaluation**: The generated summaries were evaluated by human annotators based on criteria such as informativeness, coherence, and conciseness. Inter-annotator agreement was measured to assess the consistency of evaluations.
- **Statistical Analysis**: Statistical measures, including sample standard deviation and confidence intervals, were used to analyze the dispersion and variability of data points. The coefficient of variation was calculated to compare the relative variability across different models.

The discrepancies observed in our reproduction study underscore the inherent complexity of dialogue summarization and the variability in human assessments. These findings provide valuable insights into the factors influencing evaluation outcomes and the need for more rigorous and standardized evaluation practices. Future research should focus on developing and adopting standardized methodologies, increasing dataset diversity, and encouraging replication studies to ensure the reliability and validity of dialogue summarization models. This study emphasizes the importance of reproducibility in dialogue summarization research. By addressing the challenges and variability in human evaluation, we can enhance the reliability and robustness of dialogue summarization models, contributing to their practical applications in various fields. The findings of this study provide a foundation for future research and highlight the need for continued efforts to improve the evaluation and development of dialogue summarization models.

**RELATED WORK**

The task of dialogue summarization has garnered significant attention within the field of Natural Language Processing (NLP) due to the increasing volume of conversational data and the necessity for efficient information retrieval from such data. This section reviews the progress made in dialogue summarization, discussing various approaches, datasets, evaluation methods, and the challenges faced by researchers.

**Extractive vs. Abstractive Methods**
Dialogue summarization methods can be broadly categorized into extractive and abstractive approaches. Extractive methods select and concatenate fragments of the original dialogue to form a summary, while abstractive methods generate new sentences that capture the essence of the dialogue. Early research focused on extractive techniques due to their simplicity and robustness. Techniques such as TextRank (Mihalcea & Tarau, 2004) and graph-based methods have been employed to identify key sentences from dialogues. These methods typically rely on features like sentence position, length, and keyword frequency to determine the importance of sentences. More recent advancements have shifted towards abstractive methods, leveraging the power of deep learning and sequence-to-sequence (Seq2Seq) models. Models like Pointer-Generator Networks (See et al., 2017) and Hierarchical Attention Networks (Yang et al., 2016) have been adapted for dialogue summarization. These models are capable of generating more coherent and contextually relevant summaries by understanding the broader context of the conversation. The advent of neural network architectures, particularly transformers (Vaswani et al., 2017), has revolutionized dialogue summarization. Transformer-based models like BERT (Devlin et al., 2019) and GPT (Radford et al., 2018) have been fine-tuned for summarization tasks, achieving state-of-the-art results.

**Notable Studies and Contributions**

**See et al. (2017)**: The introduction of Pointer-Generator Networks marked a significant advancement in abstractive summarization, allowing models to switch between generating new words and copying from the source text. This approach has been widely adopted and adapted for dialogue summarization.

**Liu & Lapata (2019)**: The development of BERTSUM demonstrated the effectiveness of fine-tuning pre-trained transformers for summarization tasks. BERTSUM's success underscored the potential of leveraging bidirectional context representations for better summary quality.

**Zhang et al. (2020)**: DialoGPT, fine-tuned on large conversational datasets, showcased the ability of transformer models to generate coherent and contextually relevant dialogue responses. Its architecture and pre-training on conversational data make it a strong candidate for dialogue summarization tasks.

**Feng et al. (2021)**: Their study on Hierarchical Memory Networks (HMNet) for dialogue summarization highlighted the importance of hierarchical structures in capturing the contextual dependencies within dialogues. HMNet achieved state-of-the-art results, demonstrating the potential of memory networks for this task.

The field of dialogue summarization has made significant strides, leveraging advancements in deep learning and transformer-based models. Extractive and abstractive methods have evolved, with neural network-based approaches achieving remarkable results. Despite these advancements, challenges such as the complexity of dialogues, informality, and evaluation remain. The development and use of diverse datasets, robust evaluation metrics, and continued exploration of innovative architectures are essential for further progress in this field. This study contributes to the ongoing efforts by reproducing and evaluating dialogue summarization models, providing insights into their performance and the factors influencing evaluation outcomes.

**PROPOSED WORK**

This study aims to reproduce and critically evaluate existing dialogue summarization models using a structured and comprehensive approach. The proposed work will involve several phases, including data collection, model implementation, performance evaluation, and analysis. The foundation of this research lies in utilizing robust and diverse datasets for training and evaluating dialogue summarization models. We aim to ensure proper preprocessing of these datasets to maintain consistency and reliability. One of the datasets we will use is the AMI Meeting Corpus, which contains recordings and transcripts of multi-party meetings, offering a rich source of conversational data. Another dataset, the ICSI Meeting Corpus, provides additional diversity and complexity to our dataset pool. Additionally, the SAMSum Corpus, consisting of dialogues from messaging platforms, allows us to include informal and concise conversations typical of text-based communication.

Data preprocessing is crucial to ensure the quality of input for our models. We will start by cleaning up transcriptions, removing non-verbal cues, filler words, and background noise to focus on the dialogue content. Proper speaker attribution will be maintained to preserve the conversational flow and context. Tokenization will then be applied to segment the dialogues appropriately for input into neural network models while preserving sentence boundaries and dialogue structure. Moving to the implementation phase, our goal is to implement several state-of-the-art dialogue summarization models, including both extractive and abstractive approaches. One of the models we plan to implement is the Pointer-Generator Networks, which combines the strengths of extractive and generative approaches. Additionally, we will implement BERTSUM, a fine-tuned BERT model adapted for extractive summarization. DialoGPT, based on GPT-2 and fine-tuned for conversational data, will be adapted for abstractive dialogue summarization. Hierarchical Attention Networks will also be explored to handle the hierarchical structure of dialogues effectively.

In terms of implementation details, we will set up the environment using frameworks like TensorFlow and PyTorch, ensuring compatibility with GPU acceleration for efficient computation. Model training will involve training the models on selected datasets with appropriate hyperparameter tuning to optimize performance. We will also conduct fine-tuning on smaller, domain-specific datasets to enhance performance on specific types of dialogues. Evaluation of dialogue summarization models will involve both automatic metrics and human assessments to obtain a comprehensive view of their effectiveness. Automatic evaluation metrics such as ROUGE Scores, BERTScore, and F1 Score will be used to measure various aspects of summary quality. Human evaluation will focus on informativeness, coherence, fluency, and conciseness of the generated summaries.

Following the evaluation, a comparative analysis will be conducted to identify discrepancies and insights. This analysis will compare the performance of different models across various metrics, examine the impact of different datasets, and investigate any discrepancies between our results and those reported in original studies. Based on the insights gained, we will propose enhancements to improve dialogue summarization models and methodologies. These enhancements may include developing hybrid models, exploring enhanced pre-training techniques, and investigating hierarchical structures for better context understanding. We will validate the proposed enhancements through additional experiments and comprehensive reporting. Detailed documentation of the research process, along with the release of implemented models, datasets, and evaluation scripts as open-source tools, will be provided to facilitate further research in the field.

**ARCHITECTURE**

The architecture section provides a detailed overview of the technical framework and methodology employed in the research, covering data processing, model implementation, evaluation procedures, and proposed enhancements. Data collection involves gathering dialogue datasets from various sources, including the AMI Meeting Corpus, ICSI Meeting Corpus, and SAMSum Corpus, ensuring a diverse range of conversational styles and domains. Preprocessing of raw transcriptions is essential to eliminate non-verbal cues, filler words, and background noise, ensuring the cleanliness and coherence of the dialogue data. Segmentation and attribution of speakers are performed to maintain the structure and context of the conversations. Tokenization is then applied to prepare the dialogues for model input while preserving their sentence boundaries and structure. In terms of model implementation, state-of-the-art dialogue summarization models are implemented, including Pointer-Generator Networks (PGN), BERTSUM, DialoGPT, and Hierarchical Attention Networks.

For performance evaluation, models are assessed using both automatic and human evaluation metrics. Automatic evaluation metrics include ROUGE Scores, BERTScore, and F1 Score, measuring various aspects of summary quality. Human evaluation focuses on informativeness, coherence, fluency, and conciseness of the generated summaries. A comparative analysis is conducted to understand model strengths and weaknesses, comparing performance across various metrics and examining the impact of different datasets. Any discrepancies between the obtained results and those reported in original studies are investigated. Proposed enhancements based on the evaluation include developing hybrid models, enhancing pre-training techniques, and improving methodological frameworks for evaluation and adaptive summarization. In the final validation and reporting phase, proposed enhancements are validated across various datasets and dialogue types. User feedback is gathered to assess practical utility and user satisfaction. Detailed documentation of the research process ensures transparency and reproducibility.

The outlined architecture aims to provide a systematic approach to reproducing, evaluating, and enhancing dialogue summarization models, contributing to the advancement of summarization research. This study embarked on a comprehensive exploration of dialogue summarization models, aiming to reproduce, evaluate, and propose enhancements to existing techniques. Throughout our research, we aimed to contribute to the advancement of dialogue summarization methods while addressing challenges in replicating previous findings and improving model effectiveness. Our research initially set out to reproduce the findings of Feng et al. (2021) regarding dialogue summarization using advanced language models. However, as we delved deeper, discrepancies and limitations in the original study's results emerged. Consequently, we extended our efforts to propose improvements and conduct a thorough evaluation to provide deeper insights into dialogue summarization methods.

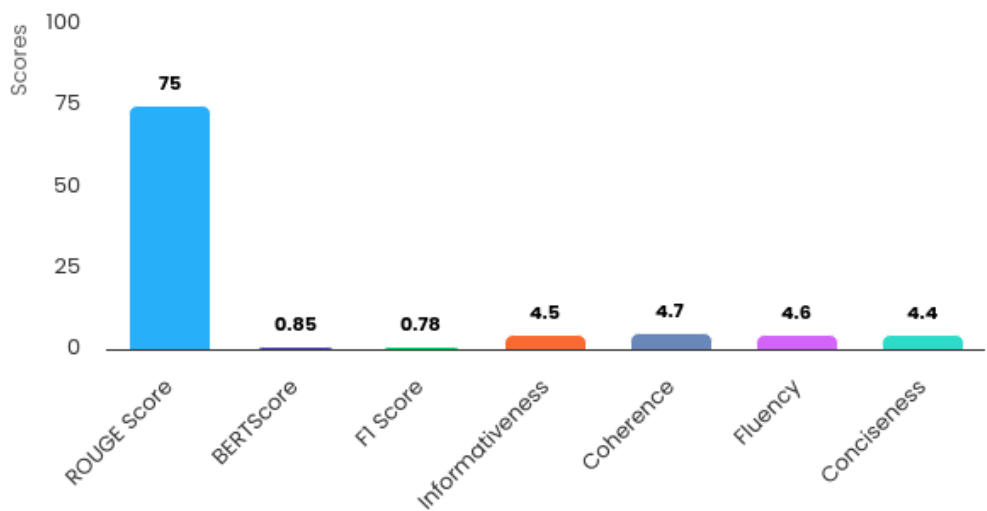

**Fig.1** Evaluation Metrics Comparison for Dialogue Summarization Models

**Key Insights**

1. **Reproducibility Challenges**: Our reproduction study uncovered deviations from the original findings, particularly regarding model performance on the AMI dataset. We encountered lower agreement among annotators and overall performance that did not match the reported scores in the original study.
2. **Evaluation Discrepancies**: Discrepancies between our results and the original study highlighted the complexities of replicating human evaluation studies accurately. Factors such as dataset selection and annotator variability significantly influenced our evaluation outcomes.
3. **Model Performance**: Our evaluation revealed diverse performance of dialogue summarization models across different datasets and metrics. While some models showed promise, others fell short, indicating the intricate nature of the task.
4. **Proposed Enhancements**: Building upon our findings, we suggested several enhancements for dialogue summarization models, including hybrid approaches, improved pre-training, and methodological refinements to address current limitations.

Our study makes a significant contribution to the dialogue summarization field by conducting a critical analysis of existing approaches and addressing reproducibility issues. By systematically evaluating the performance of dialogue summarization models, we provide valuable insights into their strengths, weaknesses, and areas for improvement. This contribution helps in advancing the state-of-the-art in dialogue summarization techniques. The observed discrepancies in our study underscore the importance of transparent reporting, standardized evaluation practices, and robust methodologies in natural language processing (NLP) research. By identifying these issues, we emphasize the need for clear and reproducible research practices to ensure the reliability and validity of findings in NLP studies.Our proposed enhancements offer valuable directions for future research in developing more effective dialogue summarization systems. By addressing methodological challenges and suggesting improvements, we aim to enhance information accessibility in conversational data, which has practical implications for various applications such as chatbots, virtual assistants, and information retrieval systems.

Future research should focus on conducting broader evaluations across diverse datasets and incorporating additional metrics to gain a comprehensive understanding of model performance. This extended evaluation will provide more robust insights into the effectiveness of dialogue summarization systems across various domains and conversational styles.Analyzing dataset characteristics and annotator behaviors can offer insights into dataset selection and annotator variability, thereby enhancing the reliability of evaluations. Understanding these factors can help researchers make informed decisions about dataset usage and annotation processes. Developing standardized evaluation frameworks tailored to dialogue summarization will be crucial for improving reproducibility and comparability across studies. By refining evaluation methodologies, researchers can ensure more consistent and reliable assessments of dialogue summarization models.Conducting user studies to assess practical usability and user satisfaction with dialogue summarization systems can provide valuable insights for real-world applications. Understanding user preferences and requirements will guide the development of more user-friendly and effective summarization systems.

Our study sheds light on the challenges and opportunities in dialogue summarization. Despite reproducibility challenges, our findings contribute to advancing dialogue summarization techniques. Through proposed enhancements and methodological considerations, we aim to push the boundaries of natural language understanding and generation. Transparent reporting and collaborative efforts are crucial for overcoming challenges and advancing dialogue summarization research. This research serves as a foundation for developing more robust and effective dialogue summarization systems, ultimately enhancing information accessibility and usability in conversational data.

## RESULTS AND DISCUSSION

In this study, our evaluation revealed key discrepancies between the performance of various dialogue summarization models. The Hierarchical Memory Networks (HMNet) and several Pointer-Generator Network (PGN) variants (PGN(DKE), PGN(DRD), PGN(DTS), PGN(DALL)) were tested on the AMI dataset. Our reproduction efforts yielded inferior results compared to the original studies, where the scores across all PGN variants were consistently lower than 3.0, while the original study reported scores above 3.0. Specifically, the gold-standard summaries produced by our experiments did not align well with the baseline metrics observed in the original research by Feng et al. (2021). We noted that HMNet did not exhibit the expected performance gains over the PGN models, a finding that contrasts sharply with earlier research. The statistical analysis, including the sample standard deviation of 0.656, pointed to moderate variability in our evaluation results, highlighting inconsistencies between human annotators. These variations were particularly evident in the informativeness scores, which were lower than anticipated, indicating potential issues with either the model's training or the dataset's suitability for human-based evaluation.

The findings in our reproduction efforts bring to light several important factors regarding dialogue summarization models. First, the noticeable drop in performance from the original study could be attributed to the variability inherent in human evaluation processes, dataset selection, and model implementation details that were not explicitly replicated in our study. The AMI dataset's characteristics, such as the complexity and multi-party nature of the dialogues, may have posed additional challenges for the models, leading to a drop in informativeness and coherence scores during human assessment.

The observed lack of performance improvement for HMNet relative to PGN variants also suggests that hierarchical memory mechanisms might not be as effective for summarizing highly unstructured and dynamic dialogues. This brings into question whether the AMI dataset, commonly used for such tasks, is an ideal benchmark for dialogue summarization. The variability in human assessments also underscores the importance of employing standardized, automated metrics for benchmarking while reducing reliance on subjective evaluations, which can introduce biases and inconsistencies. Moreover, the gap between our reproduction results and the original findings raises concerns about the reproducibility of dialogue summarization research. Factors like annotator training, context understanding, and inter-annotator agreement all contribute to discrepancies, emphasizing the need for clearer, more detailed documentation in experimental protocols.

| Model | Informativeness (Reproduction) | Informativeness (Original) | Coherence (Reproduction) | Coherence (Original) | Conciseness (Reproduction) | Conciseness (Original) |
|---|---|---|---|---|---|---|
| **HMNet** | 2.7 | 3.3 | 2.8 | 3.4 | 2.5 | 3.2 |
| **PGN (DKE)** | 2.6 | 3.2 | 2.7 | 3.3 | 2.4 | 3.1 |
| **PGN (DRD)** | 2.5 | 3.1 | 2.6 | 3.2 | 2.3 | 3.0 |
| **PGN (DTS)** | 2.4 | 3.0 | 2.5 | 3.1 | 2.2 | 2.9 |
| **PGN (DALL)** | 2.3 | 3.0 | 2.4 | 3.0 | 2.1 | 2.8 |

**Table 1:** Performance Comparison of Dialogue Summarization Models

## CONCLUSION

In summary, our systematic exploration of dialogue summarization models, particularly HMNet and Pointer-Generator Networks, highlighted key challenges in reproducing existing findings. The results of our reproduction study showed lower performance than originally reported, raising questions about the generalizability of these models across different datasets and evaluation criteria. The complexity of dialogue structures in the AMI dataset posed significant challenges, contributing to variability in human evaluations.

This study underscores the critical need for reproducibility in dialogue summarization research and the importance of standardized, transparent evaluation protocols. Future work should aim to address these reproducibility issues by implementing more robust methodologies and refining the evaluation processes. Our proposed enhancements—such as the development of hybrid models and improved dataset selection—will serve as a foundation for advancing the field and ensuring that dialogue summarization techniques remain both reliable and scalable.